\documentclass[letterpaper, 10 pt, conference]{ieeeconf}  % Comment this line out if you need a4paper

\IEEEoverridecommandlockouts                              % This command is only needed if 
\usepackage{graphics} % for pdf, bitmapped graphics files
\usepackage{epsfig} % for postscript graphics files
\usepackage{mathptmx} % assumes new font selection scheme installed
\usepackage{times} % assumes new font selection scheme installed
\usepackage{amsmath} % assumes amsmath package installed
\usepackage{amssymb}  % assumes amsmath package installed
\usepackage{booktabs} % To thicken table lines
\usepackage{hyperref}

\usepackage{xcolor}

\definecolor{ben}{rgb}{0.9,0.,0.5}

\definecolor{todo}{rgb}{0.9,0.1,0.1}

\title{\LARGE \bf
 CloudAttention: Efficient Multi-Scale Attention Scheme\\For 3D Point Cloud Learning
}

\author{Mahdi Saleh$^{1*}$ Yige Wang$^{1*}$ Nassir Navab$^{2}$ Benjamin Busam$^{1}$ Federico Tombari$^{3}$% <-this % stops a space
\thanks{*the authors contributed equally to this paper}
\thanks{**This work was partly sponsored by the German Federal Ministry for Economic Affairs and Energy (grant number: 19A19005B) through the VDA KI-Absicherung project.}% <-this % stops a space
\thanks{$^{1}$Mahdi Saleh, Yige Wang and Benjamin Busam are with the Faculty of Computer Science,
Technische Universit\"{a}t M\"{u}nchen (TUM),
85748 Garching bei M\"{u}nchen, Germany
{\tt\small m.saleh@tum.de, yige.wang@tum.de, b.busam@tum.de}}
\thanks{$^{2}$Nassir Navab is with the Faculty of Computer Science,
Technische Universit\"{a}t M\"{u}nchen (TUM),
85748 Garching bei M\"{u}nchen, Germany, and also with the Department of Computer Science, Johns Hopkins University, Baltimore,
21218 MD USA 
{\tt\small nassir.navab@tum.de}}
\thanks{$^{3}$Federico Tombari is with the Faculty of Computer Science, Technische Universit\"{a}t M\"{u}nchen (TUM),
85748 Garching bei M\"{u}nchen, Germany, and ¨
also with Google, 8002 Zurich, Switzerland
{\tt\small tombari@in.tum.de}}
}

\begin{document}

\maketitle
\thispagestyle{empty}
\pagestyle{empty}

%%%%%%%%%%%%%%%%%%%%%%%%%%%%%%%%%%%%%%%%%%%%%%%%%%%%%%%%%%%%%%%%%%%%%%%%%%%%%%%%
\begin{abstract}

Processing 3D data efficiently has always been a challenge. Spatial operations on large-scale point clouds, stored as sparse data, require extra cost. Attracted by the success of transformers, researchers are using multi-head attention for vision tasks. However, attention calculations in transformers come with quadratic complexity in the number of inputs and miss spatial intuition on sets like point clouds. We redesign set transformers in this work and incorporate them into a hierarchical framework for shape classification and part and scene segmentation. We propose our local attention unit, which captures features in a spatial neighborhood. We also compute efficient and dynamic global cross attentions by leveraging sampling and grouping at each iteration. Finally, to mitigate the non-heterogeneity of point clouds, we propose an efficient Multi-Scale Tokenization (MST), which extracts scale-invariant tokens for attention operations. The proposed hierarchical model achieves state-of-the-art shape classification in mean accuracy and yields results on par with the previous segmentation methods while requiring significantly fewer computations. Our proposed architecture predicts segmentation labels with around half the latency and parameter count of the previous most efficient method with comparable performance. The code is available at \href{https://github.com/YigeWang-WHU/CloudAttention}{https://github.com/YigeWang-WHU/CloudAttention}.

\end{abstract}

\section{INTRODUCTION}\label{section:intro}
% Point Cloud applications and difficulties(large scale, non-heterogeneity)
Point clouds are prevalent representations of 3D environments and objects which come naturally from range sensors. Point clouds are input to many robotics and augmented reality applications. Despite their wide usability, point clouds have multiple drawbacks, such as sparsity and density variations. 3D scans usually consist of millions of points represented sparsely in computers. Processing point clouds invariant to their permutation is crucial in any point cloud learning pipeline. At the same time, applying order-invariant and symmetric functions on large-scale point clouds is expensive when subtle underlying structures are relevant. Recent works suggest iterative sampling and applying order-invariant operations to point groups.

  \begin{figure}[thpb]
      \centering
      \includegraphics[width=\linewidth]{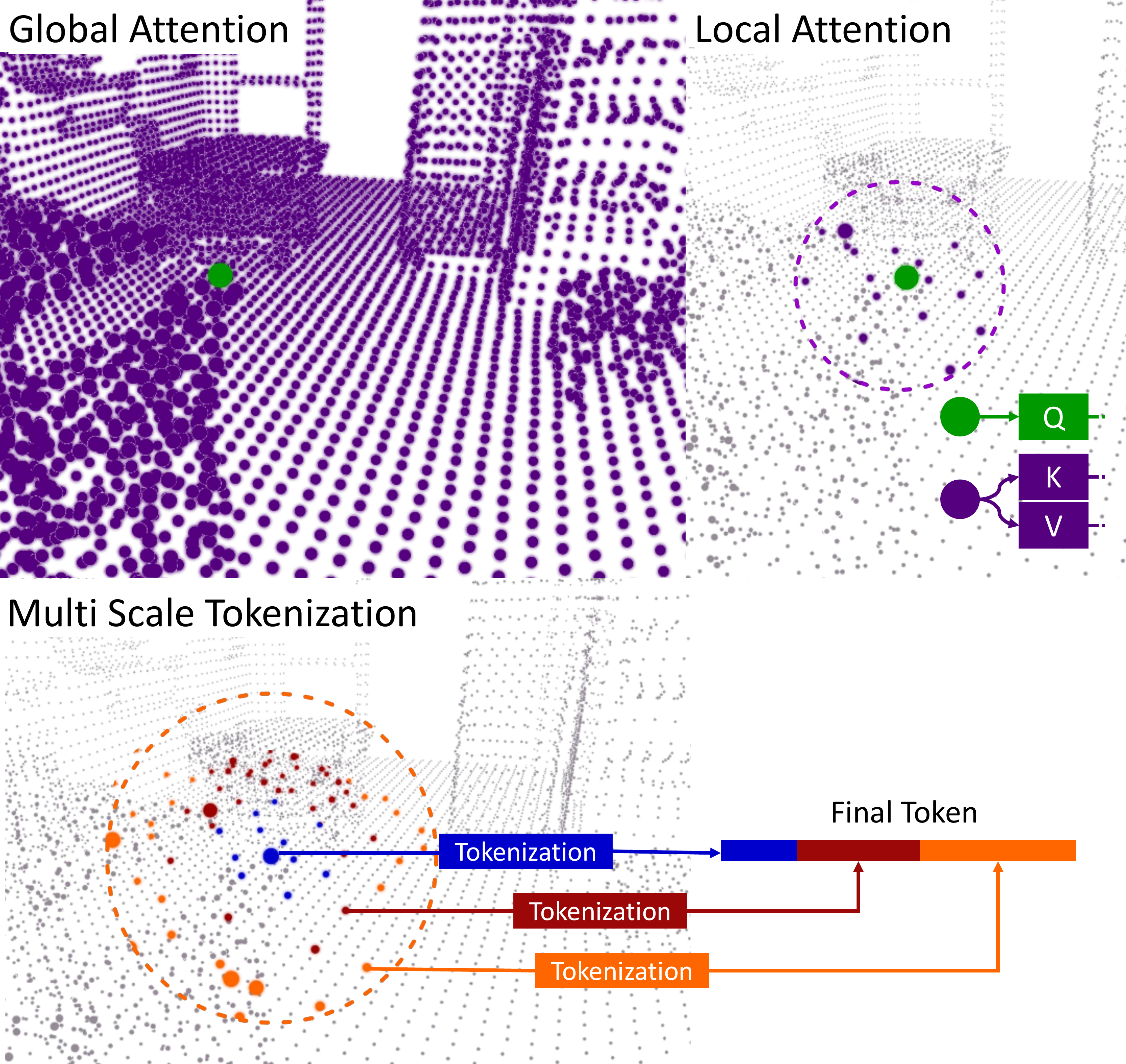}
      \caption{Top: We extract global and local attention using multi-head attention concepts, with Query Q, Key K, and Value V. Bottom: We propose an efficient multi-scale tokenization strategy to describe non-heterogeneous point clouds effectively.}
      \label{fig:teaser}
  \end{figure}
  
% Hierarchical methods PointNet++ and grouping
For instance, PointNet++~\cite{qi2017pointnet++} suggests grouping points using a ball query and applies vanilla PointNet~\cite{qi2017pointnet} on each group followed by set abstraction. While there is a hierarchical notion to PointNet++, it only focuses on the fix-sized receptive field provided by the ball query at each level. Following methods also utilize a fixed grouping using K-NN or ball query~\cite{wu2019pointconv,wang2019dynamic}. Another issue with PointNet-based methods is the extensive use of max-pooling operations, through which the gradient flow is broken and local relationships are lost.

% Transformer finding attention and expenses
Transformers~\cite{vaswani2017attention} have shown promising results in Natural Language Processing tasks and vision Transformers~\cite{dosovitskiy2020image} show the potential of utilizing Multi-head attention units for image recognition~\cite{dosovitskiy2020image}. Transformers are a good fit for sequential data since they calculate attention per node and use it for weighted pooling. Therefore, the gradient can flow to respective local neighbourhoods influenced by relationships and geometry. Transformers are applied to point sets~\cite{engel2021point,lee2019set,xu2021you} in previous works. PointTransformer~\cite{engel2021point} augments all point features by using a multi-head self-attention. In real-time use, where a huge point cloud exists, this comes with a high cost since this has quadratic complexity with respect to the number of points.

As an essential part of Transformers~\cite{dosovitskiy2020image}, self-attention is calculated to define self similarities. Using this, the nodes that are featured similarly have greater attention and are weighted higher. In previous works in point cloud processing~\cite{engel2021point} self-attention units are used to construct self-similarities across the whole input data. In this work, we conduct self-attention only to a local receptive field to focus on self-similarities with keys in the neighboring points. This way, we not only save many computations, but also can guarantee a well-distributed local attention score. 

Apart from self-attention, cross-attention is designed to capture inter-sequence attention. It is typically calculated from one query to all keys on the corresponding sequence. We use cross attention to capture global attention and feature similarities in our single frame setting. Inspired by set transformer~\cite{lee2019set}, we use raw input point cloud as the cross-domain to find global similarities to every point. By tokenizing the point cloud data and reducing the number of query tokens iteratively, we significantly reduce the number of cross-attention calculations. We call our proposed iterative cross-attention module, Global Attention Unit (GAU). In several iterations, we apply a Local Attention Unit (LAU) alternately with our GAU unit. This way, we capture local to global attention through the feature extraction stage. Figure \ref{fig:teaser}:top illustrates global and local attention.

% Multi scale tokenization mechanism
Point cloud sparsity typically varies in each sample volume. Previous methods normally perform tokenization and abstraction in a single ball or use KNN~\cite{li2018pointcnn, zhao2021point}. The ball query helps to keep the metric~\cite{qi2017pointnet++} but faces an in-balance in terms of a number of queried points. On the other hand, KNN forces the use of a fixed number of points in the neighborhood and loses metrics in varying densities~\cite{qi2017pointnet++}. This work also introduces an efficient hybrid approach where we use KNN with varying K within a ball to describe a region. Our multi-scale tokenization captures features equivariant to their level of sparsity and scale.

Combining the proposed modules reduces the computational complexity compared to previous methods and maintains a hierarchical understanding of the point cloud. The proposed efficient multi-scale feature encoder can be used in different downstream tasks such as classification and segmentation.

In summary, our contributions are the following:
\begin{itemize}
    \item We propose an efficient hierarchical architecture to learn on point clouds using iterative attention.
    \item We define self- and cross-attention modules for large-scale point clouds incorporating spatial local, and global attention mechanisms.
    \item We propose an efficient Multi-Scale Tokenization (MST) which deals with the problem of heterogeneity of sparse point clouds.
\end{itemize}

\section{RELATED WORKS} \label{section:related}
\subsection{Point Cloud Networks}
3D point clouds are unordered sparse data, and conventional CNN methods can not directly process them. Some methods quantize point clouds into regular grids and apply 3D CNNs on them~\cite{maturana2015voxnet,zhou2018voxelnet}. However, such voxel-based approaches are computationally intensive and result in high latency. 

Deep learning techniques can also process unordered sets directly using MLPs. PointNet~\cite{qi2017pointnet} builds a permutation-invariant architecture based on pointwise MLP operators. Features are pooled to form a global high-dimensional feature vector representing the whole set. PointNet++~\cite{qi2017pointnet++} extends PointNet by adding a multi-layer hierarchical procedure using furthest point sampling in each layer and grouping points by ball-query. These hierarchical operations are computationally expensive and can cause high runtime. Therefore, in large-scale applications, PointNet is usually combined with voxel-based methods~\cite{shi2020pv,zhou2018voxelnet}

While previous methods project MLP features into high-dimensional space with loose geometric intuition, graph neural networks provide a framework to extract richer local features~\cite{wang2019dynamic,saleh2020graphite}. PointCNN~\cite{li2018pointcnn} uses spatial relationship for feature aggregation of neighboring points. EdgeConv~\cite{wang2019dynamic} constructs graphs dynamically at each layer using K-NN in the embedding space. Earlier graph neural networks for point clouds are limited to node-level message passing and are impractical for large-scale point cloud applications. Following works~\cite{velivckovic2017graph, vaswani2017attention} combine node interactions and updates with edge-based attentions.

\subsection{Attention and Transformers}
Attention can be calculated in various forms to represent a weighting for every point or node. It can help focus on the most relevant parts of the input. Graph Attention Networks~\cite{velivckovic2017graph} use masked self-attentional layers to predict attention coefficients per edge, and a weighted sum is used to aggregate features per node in the graph. 

In the transformer network~\cite{vaswani2017attention} the authors propose the multi-head attention concept, which constitutes key, query, and value heads to calculate the output using scaled dot-product. The specialty of Transformers is that it is entirely based on attention operations, without convolutions or recurrent operations. Besides advances in machine translation, Transformer is applied to 2D~\cite{dosovitskiy2020image}, 3D\cite{saleh2022bending,yu2021cofinet} and spatio-temporal~\cite{ruhkamp2021attention} computer vision problems. Set transformer~\cite{lee2019set} proposes a self-attention mechanism for interactions among elements of a set based on a sparse Gaussian process. Following this, Perceiver~\cite{jaegle2021perceiver} adds cross-attention to lower-dimensional data to reduce the complexity from quadratic to linear. Our GAU design is inspired by the cross-attentions in Perceiver.

Several works have applied transformers and self-attentions to 3D point clouds. PCT~\cite{guo2020pct} is the first work to adapt the self-attention of Transformer to point cloud data. It incorporates Graph Laplacian and element-wise subtraction into the self-attention calculations. Point Transformer~\cite{zhao2021point} applies the self-attentions for K-NNs and performs a global average pooling to get the global feature vector. In another Point Transformer work~\cite{engel2021point} local features are sorted by a proposed SortNet, and self-attentions are conducted to all other points for feature augmentations. While these works suggest different self-attention concepts, they have shortcomings in costly global attention calculations.

We redesign self and cross attentions in an iterative architecture and use multi-scale tokenization to productively describe regions while saving on attention computations.
% YOGO \cite{xu2021you}

\section{METHODOLOGY}\label{section:method}
\begin{figure*}[thpb]
  \centering
  \includegraphics[width=0.95\linewidth]{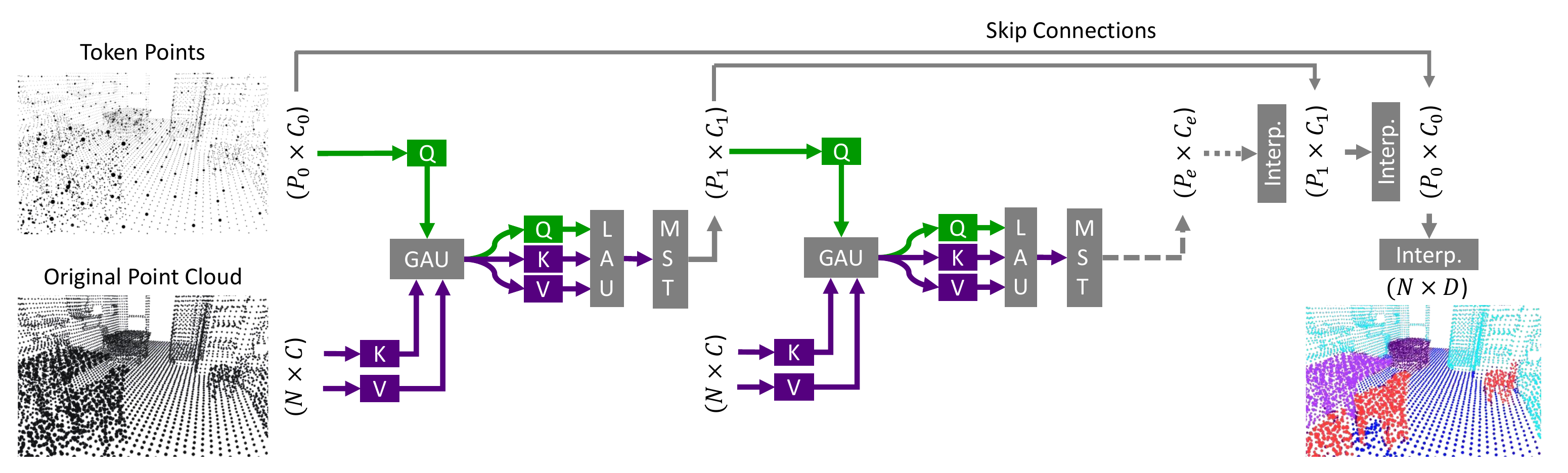}
  \caption{Our network inputs are the original point cloud and the initial tokens (left). Local and global attention are computed to encode point clouds hierarchically (center). LAUs are involved in each local neighborhood, while GAUs are applied across the tokens and the original raw point cloud. The encoded vector is passed to interpolation layers (right) to estimate the final labels per point.}
  \label{fig:overview}
\end{figure*}

This section explains our proposed method and modules one by one. Our CloudAttention extracts features from a sparse point cloud iteratively by applying local (LAU) and global attention (GAU) as well as multi-scale tokenization (MST) modules. 
Fig.~\ref{fig:overview} provides an overview of our pipeline for the semantic segmentation task.

\subsection{Notation \& Background}\label{BG}
A point cloud is a set of points, and each point is represented by its 3D coordinate and optional features, for example, color features or surface normals. Formally, given an input point cloud with $N$ points, we describe it as \(\emph{P} = {\{(p_i, f_i)\}}_{i=1}^{N} \in \mathbb{R}^{N \times d}\) where \(p_i\) is the 3D spatial coordinate of the \(i_{th}\) point and \(f_i\) its corresponding features. We stack coordinates and features vertically to form a matrix notation, i.e. \(\mathbf{p} \in \mathbb{R}^{N \times 3}\) and \(\mathbf{f} \in \mathbb{R}^{N \times (d-3)}\). For a classification task, the output is category \(C\) of the whole point cloud \(\emph{P}\), and for a segmentation task, a categorical label per every point in \(\emph{P}\).

The Multi-Head Attention (MHA) proposed in Transformer \cite{vaswani2017attention}, is an explicit form of learnable attention. It projects the input into multiple query-key-value tuples and applies the attention within each set. The attentive values across sets are aggregated and applied with a linear layer.

\subsection{Local Attention Unit (LAU)}\label{LAU}

Regular multi-head attention, is a global operation that models the relationship among all elements. However, it is not always necessary to relate elements globally. The majority of elements have negligible global relation to one another, especially regarding their low-level geometrical information. Moreover, learning global relationships in a huge point cloud is a redundant and expensive operation. Therefore, we propose a local self-attention unit (LAU) to model relationships within a vicinity. LAU is built upon self-attention concepts but only operates on a small number of neighbors. Formally, it is described as 

\begin{equation}
\begin{split}
    \text{LAU} &= S + \text{FF}(S) \\
    S &=  X + \text{MHA}(X_{i}W_Q, X_{j}W_K, X_{j}W_V) 
\end{split}
\end{equation}

\noindent where FF is simple feed-forward layer, \( X_{j} \in \Omega(X_{i},K) \) with \(\Omega(X_i,K)\) is the set of K-nearest neighbors around \(X_i\) with respect to their Euclidean distance $\|X_i-X_j\|_2$.

As LAU only operates sparsely and on a small subset of tokens instead of all tokens, the computation cost is significantly reduced compared to previous dense self-attentions strategies.

\subsection{Global Attention Unit (GAU)}\label{GAU}

Restricting attention and information exchange to small regions results in a loss of the global context, which is essential for scene segmentation tasks. Complementing our local attention, we also define a global attention function. We design a global attention unit (GAU) to incorporate global context and bring hierarchical understanding. The global attention unit is a cross-attention operation between tokens and the raw points.

\begin{equation}
    \begin{split}
        \text{GAU} &= C + \text{FF}(C) \\
        C &= T + \text{MHA}(TW_Q, PW_K, PW_V))
    \end{split}
\end{equation}

\noindent where \(T\) denotes tokens and \(P\) represents raw input point clouds. Since GAU queries from a small number of tokens to all points, the computations do not scale quadratically with the entire number of input points. Furthermore, the number of tokens is reduced iteratively in the pipeline, enabling large-scale point cloud encoding with compact operations.

\subsection{Multi-scale Tokenization (MST)}\label{MST}

Non-uniformity is intrinsic in point clouds and imposes challenges to deep learning models. Point-based learning methods largely depend on sampling densities and sparsity of the input point clouds. Nevertheless, feature estimation and tokenization in a region should depend on the geometry rather than the point density. Thus, models that counteract the influences of varying densities at low cost are more applicable to real-world settings. To this end, we propose the multi-scale tokenization (MST) technique to allow the efficient processing of input point sets.

We initially sample several centroids using farthest point sampling (FPS)\cite{qi2017pointnet++} to assure samples are distributed uniformly over the whole set. A ball query is used for each centroid to find relevant points in its vicinity. We then sort the queried points with respect to their Euclidean distance and construct a multi-scale feature within quantized distances. Figure \ref{fig:teaser} visualizes the MST operation on a point cloud region. Formally, we do multi-scale tokenization as follows:

\begin{equation}
    \begin{split}
        \text{MST} &= T_{s_1} \oplus T_{s_2} \oplus ... T_{s_N} \\
        T_{s_i} &= \Theta_{p_j \in \Omega(c_i, K_i)}(\delta((p_j - c_i) \oplus f_j))
    \end{split}
\end{equation}

\noindent where \(N\) is the total number of scales, \(T_{s_i}\) is the feature abstracted at level \(s_i\), the centroid coordinate is depicted as \(c_i\) and \(p_j\) belongs to its nearest \emph{\(K_i\)} point set \(\Omega(c_i, K_i)\). The operators \(\delta\) describe a linear layer and \(\Theta\) is a pooling operation, while \(\oplus\) denotes concatenation. We set \(K_1 < K_2 < \ldots < K_N\) to construct features at increasing scales.

MST is efficient for two reasons: 1) We use a single ball-query operation. We sort the points in the queried ball and simply form multi-scale features by tensor slicing, which omits any additional K-NN operations. 2) The linear layer \(\delta\) is shared across scales, which avoids multiple MLP usage for features abstraction.

Using a hybrid multi-scale approach, we borrow the advantages of both ball-query and K-NN methods. The method is robust to the choice of the ball radius as we tokenize a comparably large ball and rely on varying K-NNs to describe neighboring points. Contrary to previous methods that only abstract the features in a fix-sized ball, we incorporate relative distances into the output feature. Previous K-NN approaches also have an ample search space, often requiring KD-Tree search, whereas here, we do not need extra operations to find neighbors. Finally, contrary to the previous search strategies, our approach neglects irrelevant further points with the restriction imposed by the ball. We equip the representation with more flexibility by abstracting features at different scales inside the ball.

\subsection{Network \& Losses}\label{loss}

We use the encoder model for shape classification and end-to-end segmentation tasks. The tokens produced at the last stage are pooled to a single vector for global category prediction for shape classification. For segmentation, the final stage tokens are interpolated hierarchically to recover the initial resolution of the point cloud, and the interpolated features lead to per-point predictions. We use cross-entropy loss for training both classification and segmentation models with

\begin{equation}
    L_{cls} = - \sum_{c} g_c \cdot \log(p_c),
\end{equation}

\noindent where \(c\) is the category, \(g_c\) is ground truth of the \(c^{th}\) category, and \(p_c\) is the predicted probability for the \(c^{th}\) class. We only need to average over the cross-entropy losses of all points to construct the training objective for the segmentation task:

\begin{equation}
    L_{seg} =  -\frac{1}{n} \sum_{n} \sum_{c} g_c^{n} \cdot \log(p_c^n),
\end{equation}

\noindent where $n$ is the total number of points. Ground truth is given by \(g_c^{n}\) for the \(n^{th}\) sample and the \(c^{th}\) class and \(p_c^n\) represents the predicted probability, respectively.

To recover the initial resolution for segmentation, we interpolate the features at each interpolation level using inverse distance weighted average \cite{qi2017pointnet++} based on $k$ nearest neighbors.

\begin{equation}
f_c = \frac{\sum_{i=1}^{k} w_i(c)f_i}{\sum_{i=1}^{k} w_i(c)} \quad \textrm{where} \quad w_i(c) = \frac{1}{\text{d}(c, p_i)^2}
\end{equation}

\noindent where \(c\) is the position to be interpolated, \(p_i\) are its $k$ nearest neighbor coordinates of Euclidean distance $\text{d}(c, p_i)$, and \(f_i\) describes the corresponding point features.

\section{DATASETS AND METRICS}
This section introduces the datasets on which we performed our experiments, followed by their respective evaluation metrics. 

\subsection{ModelNet40}
%one
The ModelNet40~\cite{wu20153d} dataset is used for shape classification tasks. ModelNet40 consists of CAD models in 40 categories. The authors split the dataset into 9843 training models and 2468 testing models. To generate point clouds from CAD models, uniform sampling is applied to each model, along with its normal vectors. We sample 1024 points from each CAD model.

% \begin{figure}[htb!]
%   \centering
%   \includegraphics{figures/modelnet.png}
%   \caption{Data samples from ModelNet40} \label{fig:modelnet}
% \end{figure}

The evaluation metrics adopted for the shape classification task are OA (Overall Accuracy) and mAcc (mean Accuracy). OA gives the accuracy of all test data across classes, while mAcc is based on the accuracy within each class.

% , i.e.,
% \begin{equation}
%         OA = acc =  \frac{Hits}{Total} \\
%         mAcc = \frac{\sum_i acc_i}{C}  
% \end{equation}

% where \(acc_i\) is the accuracy of the \(i^{th}\) class and \(C\) is the number of classes.

\subsection{ShapeNetPart}
%one
To conduct fine-grained 3D segmentation, we use the dataset ShapeNetPart~\cite{yi2016scalable}. It contains 16881 shapes, and every shape is from one of 16 classes. In total, there exist 50 unique part classes.

% \begin{figure}[htb!]
%   \centering
%   \includegraphics[width=1.0\textwidth, height=0.4\textheight]{figures/shapenet.png}
%   \caption{ShapeNet samples (top row) and their corresponding ground truths (bottom row). } \label{fig:shapenet}
% \end{figure}
For evaluations, we use two metrics, category mIoU (cat. mIoU) and instance mIoU (ins. mIoU). For a shape \(S\) and its category \(C\), we compute the intersection over union (IoU) for each part belonging to the category \(C\) based on predictions and respective ground truth. The average IoU of all part types of category \(C\) is the mIoU for shape \(S\). By averaging mIoUs over all shapes, we also calculate ins. mIoU. Finally, for cat. mIoU, we first calculate mIoUs averaging all shapes of a given category and then average over all category mIoUs.

%\begin{equation}
%    ins. mIoU = \frac{\sum_i mIoU_i}{No.  Shapes}
%\end{equation}

%\begin{equation}
%    cat. mIoU = \frac{\sum_c mIoU_c}{C}
%\end{equation}

\subsection{3D Indoor Scene}
We also conduct experiments on large-scale scene segmentation for real scenes. We use the Stanford 3D semantic parsing dataset~\cite{armeni20163d}. It is collected by Matterport scanners from six areas, including 271 rooms. Each point belongs to one of the 13 categories. We followed the standard protocol from PointNet~\cite{qi2017pointnet} to split training and test sets. 
% \begin{figure}[htb!]
%   \centering
%   \includegraphics[width=1.0\textwidth, height=0.4\textheight]{figures/s3dis.png}
%   \caption{Three rooms from S3DIS with three views for each} \label{fig:s3dis}
% \end{figure}

Each room consists of different points and is too large to fit in memory for training. Following PointNet~\cite{qi2017pointnet}, each room is split into 1m$\times$1m blocks. During training, 4096 points are sampled from each block on the fly and are regarded as one point cloud sample. At test time, the entire scene is used for evaluation without any sampling. Three evaluation metrics are used, i.e., OA, mAcc, and mIoU. In this case, mIoU is defined as the average of IoUs over 13 classes.

\section{EXPERIMENTS AND RESULTS}
This section presents the results for every method, along with a comparison and analysis. Moreover, we perform extensive ablation studies on different aspects of the proposed model. Then, the latency and number of parameters are compared to justify the efficiency of our proposed methods. Finally, we illustrate the results of inferred scenes and object parts to provide qualitative evidence for our model's performance.

We train all models on a single NVIDIA RTX 2080 TI GPU. The batch size for all tasks is set to 16. We train our network 150 epochs for the shape classification task and 200 epochs for both part segmentation and scene segmentation tasks. Cosine annealing learning rate scheduler \cite{loshchilov2016sgdr} and Lamb optimizer \cite{you2019large} are used for optimization.

\subsection{Shape Classification}
% classification
There have been several methods benchmarking on ModelNet40~\cite{wu20153d} shape classification. Tab.~\ref{tab:complete_cls} compares our predictions with the state-of-the-art shape classification methods. As shown in the results, the proposed method achieves state-of-the-art results in terms of mAcc. The predictions of our model reach 91.4\% mean accuracy, while PointTransformer~\cite{zhao2021point} achieves 90.6\%.

% shape classification results
\begin{table}[htpb]
  \caption{Comparison with other methods for the shape classification task on ModelNet40~\cite{wu20153d}.}
  \centering
  \begin{tabular}{lrr}
    \toprule
      Method &  mAcc & OA \\
    \midrule
      3DShapeNets \cite{wu20153d}&  0.773 & 0.847  \\
      Perceiver \cite{jaegle2021perceiver} & - & 0.857 \\
      VoxNet \cite{maturana2015voxnet}&  0.83 & 0.859   \\
      %Subvolume \cite{qi2016volumetric}&  0.86 & 0.892   \\
      MVCNN \cite{su2015multi}&  - & 0.901   \\
      PointNet \cite{qi2017pointnet}&  0.862 & 0.892   \\
      Set Transformer \cite{lee2019set} &  - & 0.904   \\
      YOGO (KNN) \cite{xu2021you} & - & 0.914\\
      YOGO (Ball query) \cite{xu2021you} & - & 0.913 \\
      %PAT \cite{yang2019modeling}&  - & 0.917   \\
      PointNet++ \cite{qi2017pointnet++}&  - & 0.919   \\
      %SpecGCN \cite{wang2018local} &  - & 0.921   \\
      PointCNN \cite{li2018pointcnn} &  0.881 & 0.922   \\
      DGCNN \cite{wang2019dynamic}&  0.902 & 0.922   \\
      %PointWeb \cite{zhao2019pointweb}&  0.894 & 0.923   \\
      %SpiderCNN \cite{xu2018spidercnn}&  - & 0.924   \\
      PointConv \cite{wu2019pointconv} &  - & 0.925   \\
      %Point2Sequence \cite{liu2019point2sequence}&  0.904 & 0.926   \\
      %KPConv \cite{thomas2019kpconv}&  - & 0.929   \\
      %InterpCNN \cite{mao2019interpolated} &  - & 0.93   \\
      PointTransformer \cite{zhao2021point}& 0.906 & \textbf{0.937} \\
    \midrule
      Ours &  \textbf{0.914} & 0.929 \\
      % local self-att; MST
    \bottomrule
  \end{tabular}
  \label{tab:complete_cls}
\end{table}

\subsection{Part Segmentation}

Tab.~\ref{tab:complete_partseg} shows the performance of our method for Part Segmentation on the ShapeNetPart dataset~\cite{yi2016scalable}. The mean IoU metric suggests the similarity of the predictions with the ground truth part semantic labels. Due to saturation on this synthetic dataset, most benchmarked works report similar performance in a small band around 0.852, for instance mean IoU. We see a performance on par with most recent works in terms of mean IoU.

% part segmentation results
\begin{table}
  \caption{Comparison with other methods for Part Segmentation task on ShapeNetPart dataset~\cite{yi2016scalable}.} \label{tab:complete_partseg}
  \centering
  \begin{tabular}{lrr}
    \toprule
      Method &ins. mIoU & cat. mIoU \\
    \midrule
      PointNet \cite{qi2017pointnet} &  0.837 & 0.804   \\
      PCCN \cite{wang2018deep} &0.851 & 0.818\\
      PointNet++ \cite{qi2017pointnet++} &0.851 & 0.819 \\
      DGCNN \cite{wang2019dynamic}&  0.851 & 0.823 \\
      YOGO (Ball query) \cite{xu2021you} & 0.851 & - \\
      YOGO (KNN) \cite{xu2021you} &0.852 & - \\
      Point2Sequence \cite{liu2019point2sequence} & 0.852& -\\
      SpiderCNN \cite{xu2018spidercnn} & 0.853 & 0.817\\
      SPLATNet \cite{su2018splatnet} &0.854 & \textbf{0.837}\\
      %PointConv \cite{wu2019pointconv} &0.857&0.828\\
      %SGPN \cite{wang2018sgpn}&0.858&0.828 \\
      %PointCNN \cite{li2018pointcnn} &0.861&0.846\\
      %InterpCNN \cite{mao2019interpolated}&0.863&0.84 \\
      %KPConv \cite{thomas2019kpconv}&0.864&0.851\\
      PointTransformer \cite{zhao2021point}& \textbf{0.866} & \textbf{0.837} \\
    \midrule
    %   Ours (w/o MST) & 0.853 &  0.821 \\ 
      Ours & 0.852 & 0.830 \\
    \bottomrule
  \end{tabular}
\end{table}

\subsection{Semantic Scene Segmentation}
In this experiment, we evaluate our network performance for semantic scene segmentation. The point clouds in the Stanford 3D semantic parsing dataset~\cite{armeni20163d} are from large-scale reconstructions with millions of points. Previous methods either subsample drastically, lose fine-grained details or apply their networks on a significantly smaller voxel grid. We follow the inference procedure described in \cite{xu2021you}. In Tab.~\ref{tab:semseg_local} we compare with the methods that process voxels of size $1 \times 1 \times 1$. Our method can provide high-quality predictions compared to previous methods for this task and achieves 0.643 in terms of mAcc and 0.554 in mIoU. By conducting a hybrid multi-scale approach, we improve, for instance, over YOGO~\cite{xu2021you}, which either uses ball query or K-NN to sample and describe points.

\begin{table}
  \caption{Comparison between methods for the task of semantic scene segmentation on the Stanford 3D semantic parsing dataset~\cite{armeni20163d}.}\label{tab:semseg_local}
  \centering
  \begin{tabular}{lrrr}
    \toprule
      Method &  OA & mAcc & mIoU \\
    \midrule
      PointNet \cite{qi2017pointnet}&  - & 0.49 & 0.411 \\
      SegCloud \cite{tchapmi2017segcloud} & - & 0.574 & 0.489\\
      TangentConv \cite{tatarchenko2018tangent} & -& 0.622&0.526 \\
      YOGO (Ball query) \cite{xu2021you} & - & - &0.538 \\
      YOGO (KNN) \cite{xu2021you} & - & - & 0.54\\
      %PCCN \cite{wang2018deep} & - &0.67 & 0.583\\
    %   PointWeb \cite{zhao2019pointweb} & 0.87& 0.666&0.603 \\
    \midrule
      Ours &0.84 &\textbf{0.643} & \textbf{0.554} \\
    \bottomrule
  \end{tabular}

\end{table}

\subsection{Performance Analysis}

Besides the effectiveness of our proposed method on different tasks, our model can run at very low latency. Table~\ref{tab:performance} details latency and memory consumption compared to previous methods.
Other works' latency and GPU memory consumption are taken directly from the evaluation of YOGO~\cite{xu2021you}. All methods are evaluated with a batch size of 8. Since YOGO runs experiments on higher-end hardware, we run our model and YOGO on a single RTX 2080TI GPU and scale the numbers accordingly. Our performance with and without the MST unit is 7.76~ms and 11.43~ms, respectively. This is significantly faster than other methods, including YOGO fastest at 21.3~ms. Improving the runtime with our full model by 46.3\%, we provide faster and more accurate results than previous methods. Our memory footprint is thereby 1.7~GB and 2.87~GB for models without and with MST. 

% the number of parameters for "Ours w/o MST" is 3.21M; we have fewer parameters and lower latency
% the number of parameters for YOGO is 5.68M
% the number of parameters for "Ours w MST" is 7.79M, we have more parameters but still lower latency

\begin{table}
  \caption{Efficiency comparison with other methods for Part Segmentation on ShapeNetPart dataset~\cite{yi2016scalable}.} \label{tab:performance}
  \centering
  \begin{tabular}{lrr}
    \toprule
      Method  & Latency [ms] & GPU Memory [GB]\\
    \midrule
      SpiderCNN \cite{xu2018spidercnn} & 170.1 & 6.5 \\
      PointCNN \cite{li2018pointcnn}  & 134.2 & 2.5 \\
      DGCNN \cite{wang2019dynamic}  & 86.7 & 2.4 \\
      PointNet++ \cite{qi2017pointnet++}  & 77.7 & 2.0 \\
      RSNet \cite{huang2018recurrent}  & 73.8 & \textbf{0.8} \\
      PointNet \cite{qi2017pointnet}  & 21.4 & 1.5 \\
      %SynSpecCNN \cite{yi2017syncspeccnn}  & & & \\
      %SPLATNet \cite{su2018splatnet}  & - & - \\
      %SO-Net \cite{li2018so} & - & -  \\
      YOGO (Ball) \cite{xu2021you}  & 21.3& 1.0 \\
      YOGO (KNN) \cite{xu2021you}  & 25.6 & 0.9 \\
    \midrule
      Ours w/o MST  & \textbf{7.76} & 1.7 \\ 
      Ours w. MST   & 11.43 & 2.87 \\ 
    \bottomrule
  \end{tabular}

\end{table}

\subsection{Ablation Studies}

To understand the contribution of our pipeline components, we ablate different modules proposed in the paper. We compare the full model trained for part segmentation to models without LAU, GAU, and MST. The results are summarized in Tab.~\ref{tab:abl}.
% part segmentation ablations:
% LAU vs Non LAU
% GAU vs Non GAU
% MST vs Non MST

\begin{table}
  \caption{Ablation study on the effectiveness of local (LAU), global (GAU), and multi-scale (MST) modules. }\label{tab:abl}
  \centering
  \begin{tabular}{cccrr}
    \toprule
      LAU & GAU & MST & cat. mIoU & ins.  mIoU \\
    \midrule
      & \checkmark & & 0.817 & 0.845 \\
      \checkmark & & & 0.816 & 0.847 \\
      \checkmark & \checkmark & & 0.821 & \textbf{0.8525} \\
      \midrule
      \checkmark & \checkmark & \checkmark & \textbf{0.830} & 0.8521 \\
    \bottomrule
  \end{tabular}

\end{table}

\subsubsection{Effectiveness of LAU}
Ablating the LAU module, presented in section~\ref{LAU}, indicates that LAU improves the predictions in both cat. mIoU and ins. mIoU metrics.

%\begin{table}
%  \caption{Ablation study on the effectiveness of local (LAU) module. }\label{tab:abl_LAU}
%  \centering
%  \begin{tabular}{lrr}
%    \toprule
%      Method & cat. mIoU & ins.  mIoU \\
%    \midrule
%      Ours w/o. LAU  & 0.8167 & 0.845 \\
%      Ours w. LAU & 0.821 & 0.8525 \\
%    \bottomrule
%  \end{tabular}
%\end{table}
    
\subsubsection{Effectiveness of GAU}
Similar to LAU, we ablate the effectiveness of the GAU unit described in section~\ref{GAU}. The results show that our model benefits from Global Attention as well. 

%\begin{table}
%  \caption{Ablation study on effectiveness of global (GAU) module.}\label{tab:abl_GAU}
%  \centering
%  \begin{tabular}{lrr}
%    \toprule
%      Method  & cat. mIoU & ins. mIoU \\
%    \midrule
%      Ours w/o. GAU  & 0.816 & 0.847 \\
%      Ours w. GAU  &  0.821 & 0.8525 \\
%    \bottomrule
%  \end{tabular}
%\end{table}
    
\subsubsection{Effectiveness of MST}
Lastly, we also ablate our MST module. Compared to our full method, MST provides subtle improvements for part segmentation in cat. mIoU from 0.821 to 0.830. The comparison also verifies the effectiveness of our tokenization approach.

%\begin{table}
%  \caption{Ablation study on effectiveness of MST module.}\label{tab:abl_MST}
%  \centering
%  \begin{tabular}{lrr}
%    \toprule
%      Method & cat. mIoU & ins. mIoU \\
%    \midrule
%      Ours w/o. MST  & 0.821  &  0.8525\\
%      Ours w. MST & 0.83 & 0.8521 \\
%    \bottomrule
%  \end{tabular}
%\end{table}

\subsection{Qualitative Results}
\begin{figure}[thpb]
  \centering
  \includegraphics[width=0.8\linewidth]{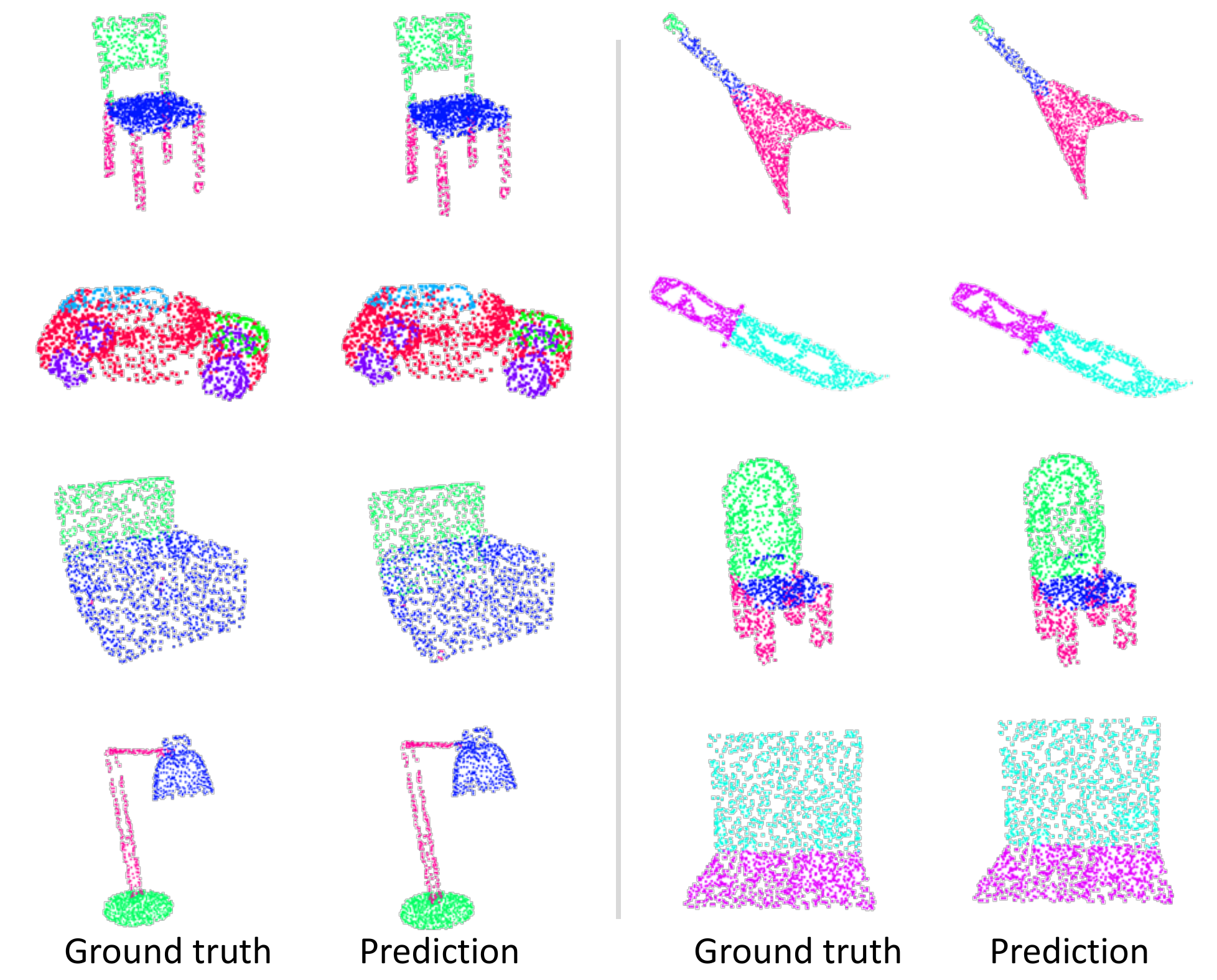}
  \caption{Qualitative results for the semantic part segmentation task on ShapeNetPart~\cite{yi2016scalable} dataset. }
  \label{fig:sem_part}
\end{figure}

Besides extensive quantitative results, we also provide qualitative results and visualization of the predictions. We show sample results from our part segmentation network in Fig.~\ref{fig:sem_part}. As illustrated, our methods predict parts classes with details very accurately. 

\begin{figure}
    \centering
    \includegraphics[width=0.8\linewidth]{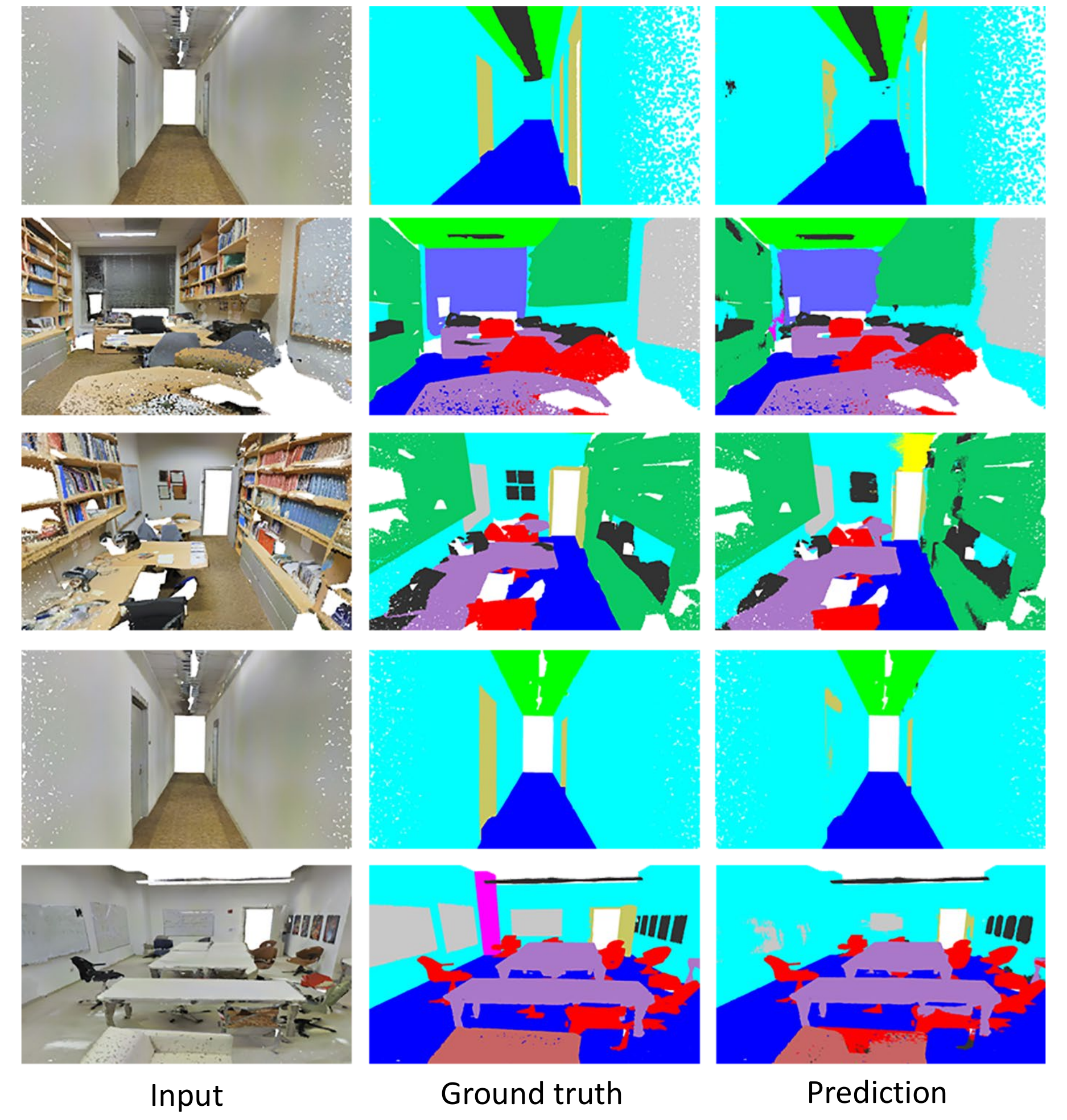}
    \caption{Qualitative results for the scene segmentation task on Stanford 3D semantic parsing dataset~\cite{armeni20163d}.} 
    \label{fig:sem_scene}
\end{figure}

Similar to part segmentation Fig.~\ref{fig:sem_scene} depicts the results for scene segmentation. The input point cloud data with embedded RGB colors is shown on the right. The comparison with ground truth shows robust predictions and supports the numerical evaluations and effectiveness of the proposed multi-scale model.

\section{CONCLUSION}
This paper presents a novel and efficient hierarchical approach to process point clouds. We design our pipeline based on attention concepts focusing on three key aspects: a) local similarly and attention help to describe local geometry with less computational burden, b) a global attention module is defined to capture global context by computing cross-attention between the sampled tokens and original cloud, and c) multi-scale tokenization describes a region by taking different levels of proximity into account. Creating a flexible and efficient point cloud processing pipeline improves standard tasks such as shape classification and semantic segmentation while improving latency. We believe that this work can pave the way to more accurate point cloud processing pipelines even on large clouds with varying densities, which is elemental for successful applications of 3D learning.

\bibliographystyle{IEEEtran}
\bibliography{bibliography}
\end{document}